\documentclass[journal]{IEEEtran}

\usepackage{times}
\usepackage{graphicx}
\usepackage{latexsym}
\usepackage{algorithm}
\usepackage{algorithmic}
\usepackage{multirow}
\usepackage{helvet}
\usepackage{courier}
\usepackage{amssymb}
\usepackage{amsmath}
\usepackage{rotating}
\usepackage{subfigure}
\usepackage{wrapfig}
\usepackage{graphicx}
\usepackage{epstopdf}

\newcommand{\argmax}{\mathop{\text{arg max}}}
\newcommand{\argmin}{\mathop{\text{arg min}}}

\newcommand{\true}{\text{\it true}}

\newcommand{\process}{\text{\it process}}
\newcommand{\Not}{not\,}

\newcommand{\taska}{\textit{task}_\alpha}

\graphicspath{{figures/}}

\usepackage{color}
\definecolor{orange}{rgb}{1,0.5,0}

\begin{document}
\title{Understanding User Instructions by Utilizing Open Knowledge for Service Robots}

\author{Dongcai Lu, Feng Wu$^\ast$,
and Xiaoping Chen}


\maketitle

\begin{abstract}
Understanding user instructions in natural language is an active
research topic in AI and robotics. Typically, natural user
instructions are high-level and can be reduced into low-level tasks
expressed in common verbs (e.g., `take', `get', `put'). For robots
understanding such instructions, one of the key challenges is to
process high-level user instructions and achieve the specified
tasks with robots' primitive actions. To address this, we propose
novel algorithms by utilizing semantic roles of common verbs
defined in semantic dictionaries and integrating multiple open
knowledge to generate task plans. Specifically, we present a new
method for matching and recovering semantics of user instructions
and a novel task planner that exploits functional knowledge of
robot's action model. To verify and evaluate our approach, we
implemented a prototype system using knowledge from several open
resources. Experiments on our system confirmed the correctness and
efficiency of our algorithms. Notably, our system has been deployed
in the KeJia robot, which participated the annual RoboCup@Home
competitions in the past three years and achieved encouragingly
high scores in the benchmark tests.
\end{abstract}


\IEEEpeerreviewmaketitle

\section{Introduction}
\label{sec:intro}

\begin{figure}[t]
\begin{minipage}{0.48\linewidth}
  \centerline{\includegraphics[width=3.5cm]{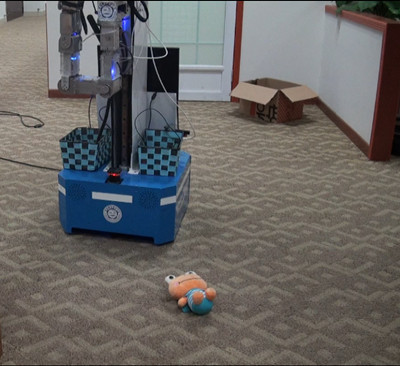}}
  \vspace{5pt}\small
  \begin{tabular}{l}
    \hline
    - Task: \\
        \hspace{.5cm}$\cdot$ Clean up toys \\
    - Steps: \\
        \hspace{.5cm}$\cdot$ Pick up toys from floor \\
        \hspace{.5cm}$\cdot$ Put toys in the toybox \\
    \hline\vspace{-5pt}
  \end{tabular}
  \centerline{(a) Clean up toys}
\end{minipage}
\begin{minipage}{.48\linewidth}
  \centerline{\includegraphics[width=3.5cm]{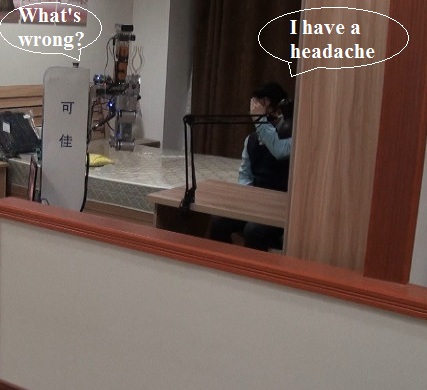}}
  \vspace{5pt}\small
  \begin{tabular}{l}
    \hline
    - Desire: \\
        \hspace{.5cm}$\cdot$ Have a headache \\
    - Actions: \\
        \hspace{.5cm}$\cdot$ Give him an aspirin \\
        \hspace{.5cm}$\cdot$ With pain medication \\
    \hline\vspace{-5pt}
  \end{tabular}
  \centerline{(b) Have a headache}
\end{minipage}
\caption{Examples of robot tasks for user instructions in natural language.}
\label{fig:problem}
\end{figure}

\IEEEPARstart{N}{owadays}, service robots can do more and more work
in our daily life, such as moving around in a house, fetching drink
or medicine for elderly people, or preparing food for a family.
They are smart and can do many complex tasks autonomously.
Nevertheless, when robots encounter user requests or tasks in an
open-ended form (e.g., through dialogs in natural language), they
often fail to response properly, not only due to possible language
processing failures but also the challenges of task planning with
incomplete knowledge. For example, as illustrated in
Figure~\ref{fig:problem}, a daily instruction ``clean up toys'' is
challenging for a robot to process if the action ``clean up'' is
{\em under-specified} and ``have a headache'' is also nontrivial
for a robot to offer help to people without grounding the helping
verb (i.e., knowing how to help). These are common tasks in
domestic scenarios and therefore it is desirable for service robots
to be able to complete such tasks given user instructions in
natural language.

Typically, user instructions are \textit{action-directed} in the
sense that the fundamental purpose of an instruction is to specify
what users want a robot to do for them. This indicates a connection
between robot understanding (i.e., knowing what the users said) and
acting (i.e., doing what the users asked). In other words,
understanding an instruction means that the robot is able to
generate a plan (i.e., sequence of actions) for the tasks specified
in the instruction
\cite{Chen2010,Dzifcak2009,Kollar2010,Nyga2012,Saxena14,Tellex2011}.
Therefore, it is crucial for the robot to have the knowledge about
the tasks and actions in order to do planning. However, some
knowledge may be missing in the instruction (e.g., ``have a
headache'' does not directly indicate that the robot should given
the user an aspirin). Consequently, the robot does not know how to
act when such instructions are presented.

Fortunately, there is more and more common knowledge available in
open resources, such as the \textit{Open Mind Indoor Common Sense}
(OMICS) database~\cite{Gupta2004},
\textit{wikihow}\footnote{\scriptsize\tt http://www.wikihow.com},
\textit{WordNet}, and many other digital dictionaries. In these
dictionaries, actions are often \textit{hierarchical} where a
high-level action is composed of several lower-level actions.
Similarly, user instructions are often specified hierarchically in
which an action is referred by an action verb or verb phrase. For
instance, ``clean up a house'' may indicate a series of subtasks
such as ``clean the table'', ``clean the floor'', etc. Therefore,
commonsense knowledge about hierarchical relations between tasks
and subtasks is useful for instruction understanding.

In our previous studies~\cite{Chen2012}, we found that a user
instruction representing a high-level task can usually be reduced
into a sequence of low-level subtasks, using hierarchical knowledge
in open resources. Furthermore, we observed that this reduction
procedure often ends up at so-called primitive tasks (i.e.,
low-level subtasks expressed in {\em common
verbs}~\cite{west1953general}). For instance, in OMICS, ``serve a
drink from fridge'' is reduced into a sequence of low-level
subtasks expressed in common verbs, such as ``go to fridge'',
``open the fridge door'', and ``take the drink'', where `go',
`open', and `take' are common verbs. Ideally, if all of the
primitive tasks in the reduction can be directly mapped into
robot's actions, the robot can simply complete the task by
executing those actions.

However, it is generally nontrivial to map primitive tasks to
robot's actions. One of the key challenges is that there is little
knowledge about {\em common verbs} in most open resources and
furthermore how they can be executed by robot with its actions. To
avoid this challenge, most of the existing approaches
\cite{Kollar2010,Nyga2012,Misra14} manually create a small set of
hand-coded robot actions for primitive tasks though their
scalability (i.e., only work for small problems) and generality
(i.e., only work for specific domains) are limited. To build a {\em
general-purpose} system for handling {\em large-scale} user
instructions, we directly tackle this challenge and consider the
follow problems: 1) how to define semantics of meanings of common
verbs, match and recover such semantics in user instructions and 2)
how to handle a large number of instructions and generate plans in
realtime using open knowledge resources.

To address these problems, we propose a novel system for service
robots to 1) process user instructions based on semantic roles of
common verbs defined in semantic dictionaries, and 2) then generate
plans for the corresponding tasks of user instructions. The
semantic roles suggest possible entities in the knowledge
representation that may be missing from or omitted in natural
instructions. In more detail, we introduce a heuristic method to
match and recover missing semantic roles from the context of user
instructions. Then, we use a planner based on {\em Answer Set
Programming} (ASP)~\cite{Gelfond1988} to exploit definitions of
common verbs in terms of semantic roles and generate a plan for the
task specified in the user instruction. By putting them together,
we built a general-purpose system for service robots that can
handle large-scale user instructions using open commonsense
knowledge.

To evaluate our approach, we conducted a corpus-based experiment on
two test sets with 11885 user tasks and 467 user desires collected
from OMICS. We also developed a prototype system and ran a case
study on a service robot in two typical domestic scenarios. Our
experimental results show substantial improvement in performance on
user instruction understanding. It is worth pointing out that the
proposed system has been successfully deployed in our
KeJia\footnote{\scriptsize
http://ai.ustc.edu.cn/en/robocup/atHome/index.php} robot, which
participates annually RoboCup@Home\footnote{\scriptsize
http://www.robocup.org/robocup-home/} competition and won the first
place once and the second place twice in the pass three years.
During the benchmark tests of the RoboCup@Home competitions, our
system is used by our robot for understanding the instructions in
English given by referees and completing the corresponding tasks.
This confirms the usefulness of our system in practice.

The remainder of this article is organized as follows.
Section~\ref{sec:problem} introduces our problem and
Section~\ref{sec:system} presents an overview of our system. Then,
Section~\ref{sec:algorithms} proposes our main algorithms, followed
by Sections~\ref{sec:semantic} and~\ref{sec:planning} describing
the two key techniques used in our algorithms. Next,
Section~\ref{sec:experiments} reports our experimental results.
Finally, Section~\ref{sec:related} briefly reviews the related work
and Section~\ref{sec:conclusions} concludes.

\section{Problem Statement}
\label{sec:problem}

\noindent We aim to building a general-purpose system so that the
robot can understand user instructions and provide service for the
user. To this end, we must solve the problem of generating a
sequence of primitive actions, which can be directly executed by a
robot, given user instructions in natural language. For example,
when a user says: ``please serve a meal for me'', the robot will
take the meal, put it on a plate, and place the plate on a table;
when a user says: ``I am thirsty'', the robot will take a drink
from the fridge and deliver it to the user. To achieve this, our
system must be able to extract a task from a user instruction in
natural language (i.e., knowing what the user said) and generate a
executable plan for the task (i.e., doing what the user asked). In
other words, natural language understanding and task planning must
be combined systematically in order to solve our problem.

In the next section, we give an overview of our system for
instruction understanding and task planning that is built by
integrating different modules.

\section{System Overview}
\label{sec:system}

\begin{figure}[t]
\centering
\includegraphics[width=1.0\columnwidth]{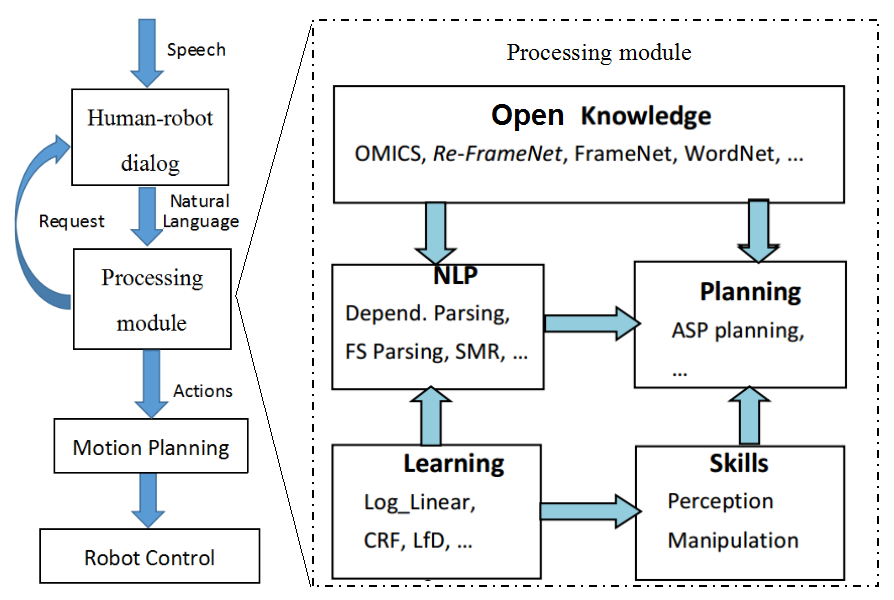}
\caption{System architecture.}\label{fig:arch}
\end{figure}

\noindent The overall architecture of our system is shown in
Figure~\ref{fig:arch}. As we can see, the human-robot dialog system
transcribes spoken utterances into text sentences and manages the
dialog with users. Each sentence in the dialog is then transferred
to the processing module, which generates a sequence of primitive
actions for the task expressed in natural language. After that, a
sequence of commands corresponding to each primitive action is
computed by the Motion Planning module. Finally, the commands are
executed by the Robot Control module.

Here, we focus on the {\em Processing module} that takes a text
sentence as its input and outputs a sequence of primitive actions
that are executable by the robot. The main components of our
Processing module are described in detail as follows.

\subsection{Open Knowledge}

\noindent As shown in Figure~\ref{fig:arch}, we use open knowledge
both for {\em Natural Language Processing} (NLP) and task planning.
The \textit{Open Knowledge} considered in our system includes
OMICS, \textit{FrameNet}, and \textit{Re-FrameNet} as introduced
below.

OMICS~\cite{Gupta2004} is an extensive collection of knowledge for
indoor service robots gathered from internet users. Currently, it
contains 48 tables capturing different sorts of knowledge, among
which the \textit{Help} and \textit{Tasks/Steps} tables are most
useful for our system. Each tuple of the \textit{Help} table maps a
user desire to a task that may meet the desire (e.g., $\langle$
``feel thirsty'', ``by offering drink'' $\rangle$). Each tuple of
the \textit{Tasks/Steps} table decomposes a task into several steps
(e.g., $\langle$ ``serve a drink'', 0. ``get a glass'', 1. ``get a
bottle'', 2. ``fill class from bottle'', 3. ``give class to
person'' $\rangle$). Given this, OMICS offers useful knowledge
about hierarchism of naturalistic instructions, where a high-level
user request (e.g., ``serve a drink'') can be reduced to
lower-level tasks (e.g., ``get a glass'', $\cdots$). Another
feature of OMICS is that elements of any tuple in an OMICS table
are semantically related according to a predefined template. This
facilitates the semantic interpretation of the OMICS tuples.

\textit{FrameNet}\footnote{\scriptsize\tt
https://framenet.icsi.berkeley.edu/fndrupal/} is a digital
dictionary providing rich semantic information for action verbs. It
groups action verbs into {\em Frames} and specifies word
definitions in terms of semantic roles called \textit{Frame
Elements} (FEs) for each Frame~\cite{Baker1998}. Although the
connections between an action verb and its semantic roles are
useful for resolving under-specification of naturalistic
instructions, this knowledge cannot be directly used by robots
since it is not formalized in \textit{FrameNet}. To overcome this
difficulty, we developed \textit{Re-FrameNet}
--- a formalized version
of \textit{FrameNet} by rewriting part of \textit{FrameNet}
knowledge in a formal meta-language.

Specifically, in \textit{Re-FrameNet}\footnote{\scriptsize\tt
http://ai.ustc.edu.cn/en/research/reframenet.php}, a Frame of
\textit{FrameNet} is formalized as a meta-task and re-defined by a
set of precondition, postcondition, invariant, and/or steps over
semantic roles of the meta-task. In the definition, FEs (i.e.,
semantic roles) such as \textit{Theme, Source}, and \textit{Goal}
of the Frame are taken as meta-variables. Therefore, the definition
of a meta-task specifies the common semantic structure of action
verbs in the corresponding Frame. For example, the meta-task
\textit{put-Placing} is defined as:
\begin{small}
\begin{align*}
&(\ \textbf{define } (\ \textbf{meta-task } \textrm{put-Placing}\\
&\hspace{.5cm} (\ \textbf{:parameters } \textit{?Agent ?Theme ?Source ?Goal}))\\
&\hspace{.5cm} (\ \textbf{:precondition  } \textit{...} )\\
&\hspace{.5cm} (\ \textbf{:postcondition  } \textit{...} )\\
&\hspace{.5cm} (\ \textbf{:invariant  } \textit{...} )\ )
\end{align*}
\end{small}
where all action verbs in Frame {\em Placing} (e.g., \textit{lay,
heap, deposit}) share the same definition. When a robot tries to
plan with \textit{put-Placing} as its action verb (verb sense) for
an instruction, our NLP components will try to extract appropriate
entities for every semantic roles specified in the definition of
meta-task \textit{put-Placing} (See Section~\ref{sec:semantic} for
more detail).

It is worth noting that \textit{common verbs} are normally not
explained in the aforementioned open resources because most of them
belong to the so-called {\em General Service List} (GSL)
--- a list of roughly 2000 most frequent English
words~\cite{west1953general}. The GSL is taken as the defining
vocabulary of dictionaries such as the Longman Dictionary of
Contemporary English, based on the notion that words should be
defined using ``terms less abstruse than the word that is to be
explained'' ~\cite{bogaards1996dictionaries}. As a result, there
are few definitions of the GSL verbs in OMICS or other digital
dictionaries.

\subsection{NLP Module}

\noindent This module maps user instruction in natural language $I$
to the OMICS tables, which contains tuple $\langle task, steps
\rangle$ for task-oriented instructions or tuple $\langle desire,
task \rangle$ for desire-oriented instructions (See
Section~\ref{sec:algorithms} for more detail). The output is a
logical form $L$ to the \textit{Planning} module, containing a
frame-semantic representation as:
\begin{small}
\begin{align*}
&(\textbf{meta-task } \textrm{take-Taking } \\
&\hspace{.5cm} (\ : \textbf{parameters } \texttt{food fridge} )\ )
\end{align*}
\end{small}
Specifically, interpreting $I$ to $L$ is done in three steps: 1)
{\em dependency parsing} that analyzes the dependencies of each
word in a sentence, 2) {\em frame-semantic parsing} that identifies
the verb's frame, and 3) {\em semantic matching and recovering}
that fills the semantic roles for a given frame. In
Section~\ref{sec:semantic}, each step will be described in detail.

\subsection{Planning Module}

\noindent The \textit{Planning} module takes the logical form of
user instruction $L$, online knowledge base (e.g.,
\textit{Re-FrameNet, WordNet, FrameNet}), domain knowledge, and
robot's skills as the inputs. The output of the \textit{Planning}
module is a high-level plan for the motion planning module.

We employ both \textit{global} and \textit{local} planners in the
{\em Planning} module. The global planner searches through the
whole knowledge of task decomposition in OMICS to generate a plan.
However, most of tasks is OMICS cannot be decomposed into robot's
primitive actions because many steps in OMICS are referred by
common verbs, for which OMICS does not contain decomposition
knowledge. For example, verbs such as \textit{take, place, put,
get}, and \textit{turn} frequently occur in task steps but there is
no knowledge in OMICS about how to execute them by the robot.
Therefore, a local planner based on ASP is used for planning based
on merely the instruction itself.

Note that the local planner is incapable of generating a plan for
under-specification terms in an instruction. Therefore, common
verbs referred by the instruction must be specified first in order
to generate a plan. Fortunately, semantic dictionaries such as
\textit{FrameNet} provide rich knowledge about common verbs. In
{\em Re-FrameNet}, we reorganize the definition of an action verb
by a set of precondition, postcondition, and invariant over
semantic roles of the action (a.k.a., the functional definition of
action). Given this, a planner based on ASP can plan actions for
the instruction using the formalized functional definition of an
action. Section~\ref{sec:planning} will give more detail about our
planning method.

\subsection{Skills and Action Model}

\noindent For a robot, we define an {\em Action Model} to specify
its skills. Specifically, an Action Model consists of several
primitive actions. Each primitive action $a$ is defined by a set of
precondition, postcondition and invariant, similar to the
definition of a common verb in \textit{Re-FrameNet}. In other
words, they specify conditions under which $a$ can be executed,
conditions that hold when $a$ finishes, and conditions that must be
satisfied during the execution of $a$ respectively. Indeed, a
primitive action is the formal specification of a robot skill. As
we will show later sections, the Action Model is useful for our
system to generate a plan that is executable by the robot.

\subsection{Learning Module}

\noindent In this module, methods such as {\em log\_linear}, {\em
Conditional Random Field} (CRF), {\em Learn from Demonstration}
(LfD) are used to learn robot's low-level skills. Intuitively, the
more skills a robot possesses, the more capable it is. For example,
unless a robot knows how to pour water to a cup, it cannot finish
the high-level task such as ``make a coffee'' (with the task-step
tuple $\langle$ ``make a coffee'', 0. ``put hot water in a cup'',
1. ``pour the coffee'' $\rangle$). In this paper, we assume that
our robot has all necessary low-level skills to complete a task
specified by user instructions though most of the skills must be
learned one by one in practice. The learning methods for robot
skills are interesting but beyond the scope of this article.

After introducing our system as a whole, we describe our main
algorithms for instruction understanding next.

\section{Understanding User Instructions}
\label{sec:algorithms}

\begin{algorithm}[t]\small
\caption{SolveTask(task \textit{t}, ActionModel \textit{AM})}
    \begin{algorithmic}[1]
        \STATE $gSeen := \varnothing$ /* prevent infinite recursive loop when exploratory searching itself */
        \STATE initiate ${worldmodel}$ and ${plans}$
        \IF{$\textit{t} \in gSeen$}
        \RETURN $null$
        \ENDIF
        \STATE $gSeen = gSeen \cup \textit{t}$
        \STATE $subTasks$ = FindSubTasks(t) /* find subtasks of task $t$ from the {\em Tasks/Steps} table in OMICS */
        \FOR{each task \textit{s} in $subTasks$ }
        \IF{  GeneratePlans(\textit{s}, \textit{AM}) $= null $}
        \STATE $FoundEqualTask = False$
        \WHILE{there is a new $t^{'}$ from the {\em Tasks/Steps} table semantically equivalent to \textit{s}}
        \IF{ SolveTask( $t^{'}$,$AM$ ) $\neq null$}
        \STATE $FoundEqualTask = True$
        \STATE $plans$.append(SolveTask$(t^{'},AM )$)
        \STATE $wordmodel$ = simulator$(wordmodel, plans)$
        \STATE \textbf{break}
        \ENDIF
        \ENDWHILE
        \IF{ $FoundEqualTask = False$ }
        \RETURN $null$
        \ENDIF
        \ELSE
        \STATE $plans$.append(GeneratePlans(\textit{s},AM))
        \STATE $wordmodel$ = simulator$(wordmodel, plans)$
        \ENDIF
        \ENDFOR
        /*successfully planned*/
        \RETURN $plans$ /* all steps have been solved. */
    \end{algorithmic}
\end{algorithm}

\begin{algorithm}[t]\small
\caption{SolveHelp(desire ${t}$, ActionModel ${AM}$)}
    \begin{algorithmic}[1]
        \STATE $AllHelps$ := FindHelpsMaptoDesire(desire ${t}$) \\ /* find all help tasks mapped to desire t*/
        \FOR{each help task ${s}$ in $AllHelps$ }
        \IF{ GeneratePlans(${s}$, ${AM}$) $= null $}
        \FOR{ task $gs$ in Tasks/Steps Table}
        \IF{ $gs$ semantically equivalent to $s$}
        \RETURN SolveTask($gs$,AM)
        \ENDIF
        \ENDFOR
        \ELSE
        \RETURN GeneratePlans(${s}$, ${AM}$)
        \ENDIF
        \ENDFOR
        \RETURN $null$
    \end{algorithmic}
\end{algorithm}

\begin{algorithm}[t]\small
\caption{GeneratePlans(task ${t}$, ActionModel ${AM}$)}
    \begin{algorithmic}[1]
        \STATE /* generate a plan for low-level task ${t}$*/
        \STATE $sem$ := SemanticMatchAndRecover(${t}$)
        \IF { $sem.frame = NULL$}
        \RETURN $null$
        \ENDIF
        \IF { $sem.frame \in {AM}$ }
        \RETURN $sem.frame(sem.parameters)$
        \ELSE
        \STATE $gRFN$ := FindRFNBySem($sem.frame$) /* find the definition of sem.frame in Re-FrameNet */
        \STATE $Res$ = solver($gRFN, sem, AM$) /* compute a plan by inputting rules of gRFN, sem and AM. */
        \IF { $Res \neq NULL$}
        \RETURN $Res$
        \ELSE
        \RETURN $null$
        \ENDIF
        \ENDIF
    \end{algorithmic}
\end{algorithm}

\noindent There are two types of user instructions that we consider
in this article: 1) {\em task-oriented} instruction (e.g., ``serve
a meal'') and 2) {\em desire-oriented} instruction (e.g., ``I am
thirsty''). In OMICS, a task-oriented instruction is represented as
tuple $\langle t, s \rangle$, where $s = \langle s_1, s_2, \cdots,
s_n \rangle$ is a sequence of the $n$ steps to complete the task
$t$. For example, given task $t =$ ``serve a meal'', a sequence of
steps may be $s = \langle s_1:$ ``take the meal'', $s_2:$ ``put it
on a plate'', $s_3:$ ``place the plate on a table''$\rangle$.
Similarly, a desire-oriented instruction is represented as tuple
$\langle d, t \rangle$, where $t$ is the task corresponding to the
user desire $d$. For instance, given user desire $d =$ ``I am
thirsty'', the task for a robot may be $t = $ ``serve a drink''.
Indeed, in most of the domestic scenarios, a user instruction is
usually either task-oriented or desire-oriented. Now, we turn to
our algorithms for generating a plan for these two types of user
instructions respectively.

Algorithm 1 is used to process task-oriented instructions by
utilizing the {\em Tasks/Steps} table in OMICS. The input is a
naturally expressed task $t$ and the robot's action model $AM$ and
the output is a sequence of primitive actions $plans$.
Specifically, it first finds all subtasks of task $t$ from the {\em
Tasks/Steps} table of OMICS. Then, it tries to generate a plan
(i.e., a sequence of primitive actions) for each subtask. If a plan
is successfully generated, the plan is added to the plan list
$plans$ and the simulator advances to the next subtask. Otherwise,
it searches the {\em Tasks/Steps} table of OMICS again for all {\em
Semantically Equivalent} (SE) tasks\footnote{For example, the tasks
of ``give someone an object'' and ``take an object to someone'' are
semantically equivalent.} of that subtask until one of the SE tasks
is successfully planned. If there is no SE task or none of the SE
tasks can be successfully planned, a $null$ is returned to indicate
the failure of task planning. After all subtasks are successfully
planned, $plans$ are returned and executed by the robot.

Algorithm 2 is used to process desire-oriented instructions by
utilizing the {\em Help} table in OMICS. Similarly, the input is a
desire and an action model and the output is a plan. Specifically,
it first finds a list of help tasks offering the corresponding help
when given a desire. Then, it tries to plan for each of the help
tasks by checking whether the help task can be successfully planned
with a sequence of primitive actions. If so, the resulting plan is
returned. Otherwise, it searches the {\em Tasks/Steps} table in
OMICS for a SE task of the help task and calls Algorithm 1 to
generate the plan.

Notice that both Algorithms 1 and 2 depend on Algorithm 3 to
generate a plan for a low-level task $t$. In Algorithm 3, it first
performs semantic role matching and recovering for task $t$ and
outputs a frame and its roles. If no verb frame is identified, the
process terminates with $null$ as no plan can be generated. If the
frame is a primitive action, this frame plus its roles are
returned. Otherwise, the frame is evoked by common verbs. In this
case, it first finds the definition of $sem.frame$ in {\em
Re-FrameNet} and translate it to a set of rules. After that, it
computes a plan based on the rules of $gRFN$, the frame $sem$, and
the action model $AM$.

Now, the key procedures in Algorithm 3 are: 1) how to do semantic
role matching and recovering given a task expressed in natural
language; 2) how to compute a plan given a set of rules, a frame,
and an action model. The details about the two procedures are
described in Sections~\ref{sec:semantic} and~\ref{sec:planning}
respectively.

\section{Semantic Matching and Recovering}
\label{sec:semantic}

\begin{table*}[t]
\begin{center}
\caption{Data collected from FrameNet and annotated from
OMICS.}\label{tab:data}\vspace{5pt} \centering
\begin{tabular}{llllll}
\hline
{\bf Data}& {\bf Size} &{\bf Examples} & {\bf Verb}& {\bf LU} & {\bf Frame}\\
\hline
\multirow{3}*{FrameNet} & \multirow{3}*{191740} & i want to bring your daughter up to the prison & bring &bring.v & Bringing \\
&& i was visited by one of the king 's most important officials & visited & visit.v & Arriving \\
&& cutting his wrist and jumping from a third-floor window & cutting & cut.v & Cause\_harm \\
\hline
\multirow{2}*{OMICS} & \multirow{2}*{1100} & remove objects from surface & remove & remove.v & Removing \\
&& complete the dance together & complete & complete.v & Activity\_finish \\
\hline
\end{tabular}
\end{center}
\end{table*}

\noindent We propose a three-phase procedure to translate a user
instruction expressed in natural language into the internal
representation, which can be handled by our planner. Firstly, a
probabilistic syntactic parser is used to retrieve the dependencies
of the instruction. Secondly, the frame of sentence's verb is
identified by frame-semantic parsing. Here, without loss of
generality, we assume that each instruction represents just a
single task (verb). Thirdly, the semantic roles of the frame are
recovered and filled as much as possible with the matched entities
appeared in the instruction or its sentential context, represented
as a meta-task in \textit{Re-FrameNet}. More details about our
three-phase procedure is described below.

\subsection{Dependency Parsing}

\begin{figure}[t]
\centering
\includegraphics[width=0.68\columnwidth]{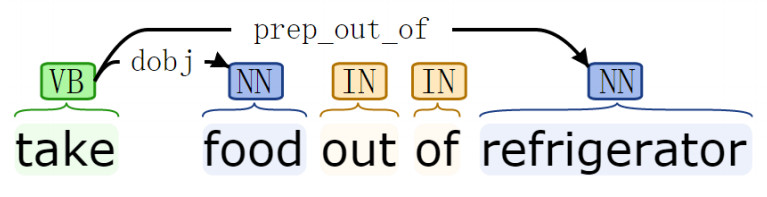}
\caption{Stanford typed dependencies of ``\textit{take food out of fridge}''.}\label{fig:syn}
\end{figure}

\noindent We use the Stanford parser~\cite{DeMarneffe2006} in the
first phase, which produces the Stanford-typed dependencies between
words in a sentence. These dependencies indicate the grammatical
relations between words in terms of the name of relation, governor,
and dependence~\cite{DeMarneffe2008}. Figure~\ref{fig:syn}
illustrates the parsing of a sentence ``take food out of
refrigerator''. The edge of the type \textit{dobj} denotes that the
noun ``food'' is the direct object of the verb ``take''. The verb
``take'' also governs the noun ``refrigerator'' via the typed
dependency \textit{prep\_out\_of}. Since the typed dependency
between a verb and a noun reveals their semantic-role relation, the
syntactic structure of an instruction is used for our semantic role
matching and recovering.

\subsection{Frame Semantic Parsing}

\noindent Given that a verb varies in different senses, an
instruction may represent different meanings and therefore can be
mapped to different frames in \textit{FrameNet}. For instance, the
verb ``take'' can represent the Frame \textit{Bring} or
\textit{Removing} under different contexts. The Stanford parser
does not disambiguate verb senses. Therefore, we propose a {\em
Frame Semantic Parsing} method to map a verb to a unique Frame.
Specifically, we define a frame identification model and train the
model with sets of data from \textit{FrameNet} and OMICS as below.

\subsubsection{Model}

Given a sentence $\mathbf{x}=\langle x_1,\dots,x_n\rangle$ with
frame-evoking verb $v$, we seek the most likely Frame $f^*$ in the
frame identification stage. Let $\mathcal{F}$ be the set of
candidate Frames for $v$, $\mathcal{L}$ the set of verbs found in
the {\em FrameNet} annotations, and $\mathcal{L}_f \subseteq
\mathcal{L}$ the subset of verbs annotated by evoking the Frame
$f$. The frame identification can be formalized by the following
prediction rule:
\[
    f^* = \argmax\limits_{f\in \mathcal{F}}\sum_{l\in\mathcal{L}_f} p(f,l|v,\mathbf{x})
\]
For $f\in\mathcal{F}$ and $l\in\mathcal{L}_f$, a conditional
log-linear model is used to model the probability
$p(f,l|v,\mathbf{x};\theta)$:
\[
    p(f,l|v,\mathbf{x};\theta) =
    \frac{\exp[\theta\cdot \Phi(f,l,v,\mathbf{x})]}{\sum_{f^\prime\in\mathcal{F}}
        \sum_{l^\prime\in\mathcal{L}_{f^\prime}}
        \exp[\theta\cdot \Phi(f^\prime,l^\prime,v,\mathbf{x})]}
\]
where $\theta\cdot \Phi(f,l,v,\mathbf{x})$ is the inner product
$\sum_{i=1}^M \theta_i \times \Phi_i(f,l,v,\mathbf{x})$ and
$\theta$ is the parameter vector over the feature function $\Phi$
with $M$ dimension.

Generally, the feature function allows for a variety of (possibly
overlapping) features. A feature $\Phi_i$ may relate a frame $f$ to
a verb $v$, representing a lexical-semantic relationship.

\subsubsection{Data}

Our training and test sets come from {\em FrameNet} lexicon and
OMICS. The {\em FrameNet} lexicon is a taxonomy of manually
identified general-purpose Frames in English. Listed in the lexicon
with each Frame are several lemmas (with part of speech) that can
denote the Frame or some aspect of it --- these are often called
{\em Lexical Units} (LUs). Table~\ref{tab:data} shows some examples
of our training and test sets.

\subsubsection{Training}

Given the training subset of the data in the form ${ \langle
 {x^{j}, v^{j}, f^{j}, s^{j}} \rangle}_{j=1}^{N}$ where $N$
is the number of sentences, we discriminatively train the frame
identification model by maximizing the following log-likelihood
function:
\[
    \max_{\theta}\sum_{j=1}^{N}log\sum_{l\in\mathcal{L}_f^{j}} p(f^{j},l|v^{j},\mathbf{x}).
\]
Specifically, we optimize it using a distributed version of
gradient ascent algorithm with initial value $\vec \theta$ as:
\begin{align*}\small
&\textbf{for } k = 0..D-1 \\
&\hspace{0.5cm} \textbf{for } i = 1..M \\
&\hspace{1cm} \theta_i = \theta_i + \alpha \frac{\partial
\sum_{j=1}^{N}log\sum_{l\in\mathcal{L}_f^{j}} p(f^{j},l|v^{j},\mathbf{x})}{\partial \theta}
\end{align*}
where $D$ is a parameter that controls the number of passes over
the training data, $M$ is the number of features, and $N$ is the
total size of our training set.

Note that the computational complexity of the algorithm above is
$O(D \times M \times N)$. When the number of features is large, it
will be costly to train our model sequentially. In order to update
the parameter of a feature $f$ faster, we consider $N_f$ training
examples that contains only $f$ instead of $N$. Hence, the
computational complexity becomes $O(D \times M \times N_f)$, where
$N_f$ is usually much smaller than $N$.

\subsection{Roles Matching and Recovering}

\noindent After the Frame for the meta-task achieved from
\textit{Re-FrameNet} is identified, the semantic roles of the
meta-task must be filled with the corresponding entities (expressed
by nouns) in the sentence or from its sentential context. In
Figure~\ref{fig:fflow}, given steps $\textbf{s} = \langle
s_1,...,s_n \rangle$ and Frames of each step $\textbf{f} = \langle
f_1,...,f_n \rangle$, we match and recover missing semantic roles
of each Frame $\textbf{r} = \langle r_1,...,r_n \rangle$, where
$r_i = \langle r_{i1},...,r_{ik_i} \rangle$.

\begin{figure}[htbp]
\centering
\begin{tabular}{l}
$s_1 : frame(f_1), role(r_{11}), role(r_{12}), ..., role(r_{1k_1})$ \\
$s_2 : frame(f_2), role(r_{21}), role(r_{22}), ..., role(r_{2k_2})$ \\
$\cdots$\\
$s_n : frame(f_n), role(r_{n1}), role(r_{n2}), ..., role(r_{nk_n})$
\end{tabular}
\caption{Formalization description of instruction flow.}\label{fig:fflow}
\end{figure}

Take the flow of instructions $\langle$ step 1: ``go to fridge'';
step 2: ``open the fridge door''; step 3: ``take the beer''; step
4: ``close the fridge door'' $\rangle$ for example. The third
instruction (i.e., step 3) is identified as the meta-task
\textit{take-Taking}, whose semantic roles in \textit{Re-FrameNet}
include \textit{Agent}, \textit{Theme}, and \textit{Source}.
However, this instruction only explicitly specifies the role
\textit{Theme (the beer)}, while the others are missing from it.
Note that the semantic role \textit{Source} can be recovered and
matched with the entity \textit{fridge} in the sentential context
of this instruction. Therefore, the challenge of our third phase
lies in the recovering of missing semantic roles.

\begin{table}[t]
\begin{center}
\caption{Heuristic rules for semantic role filling within sentence.}\label{tab:rules}
\centering
\begin{tabular}{p{2.9cm}p{2.2cm}p{2.0cm}}
\hline
\textbf{Meta-task} & \textbf{Dependency Type} & \textbf{Semantic Role} \\
\hline
\textit{put-Placing} & \textit{dobj} & \textit{Theme}\\
\textit{put-Placing} & \textit{prep\_in} & \textit{Goal}\\
\textit{take-Removing} & \textit{dobj} & \textit{Theme}\\
\textit{take-Removing} & \textit{prep\_from} & \textit{Source}\\
\textit{dry-Cause\_to\_be\_dry} & \textit{dobj} & \textit{Dryee}\\
\textit{deliver-Delivery} & \textit{prep\_to} & \textit{Recipient}\\
$\cdots$\\
\hline
\end{tabular}
\end{center}
\end{table}

To address this challenge, we borrow ideas from the ``last
objects'' method~\cite{Cantrell2010} and propose the following
method:
\begin{enumerate}
\item For any semantic role $r$ that is defined in
    \textit{Re-FrameNet} but missing from a sentence $s$, an
    entity $e$ that matches $r$ according to the definition and
    has less sentential distance from $s$ is preferable to be
    the value of $r$. Here, the sentential distance between $e$
    and $r$ is defined as $(n-m)$, if $e$ and $r$ appear in the
    $m$-th and $n$-th sentences in the same sentence flow
    respectively, with m $\leq$ n. For $1 \leq k \leq n$, it is
    formalized as: $r_{ki} = \argmin_{e \in r_l}(k-l)$, if
    $r_{ki}$ is missing and $e$ matches $r_{ki}$.
\item If a semantic role $r$ cannot be recovered through 1), it
    is assumed that (the value of) $r$ is unspecified in the
    sense that any entity satisfying the \textit{Re-FrameNet}
    definition of $r$ is a default value of $r$ under the given
    context \footnote{Some of unspecified roles should be
    identified by
    grounding~\cite{Kollar2013,Tellex2011,Varadarajan2012,Williams2013},
    which is beyond the scope of this article.}. For instance,
    the \textit{Source} role of single sentence ``put beverage
    in the fridge'' is unspecified and thus any entity in the
    class \textit{beverage} can be taken as the value of
    \textit{Source} under the context of this sentence.
    Obviously, all missing semantic roles of the first sentence
    in a flow of instructions are unspecified. In fact, given a
    context, not all of the semantic roles specified in
    \textit{FrameNet} or \textit{Re-FrameNet} are necessary for
    naturalistic language instruction understanding and task
    planning.
\end{enumerate}

In general, we divide semantic matching and recovering into two
cases. The first case is for zero sentential distance, i.e.,
recovering semantic roles based on the instruction itself.
Table~\ref{tab:rules} shows some heuristic rules for this case,
each assigning a noun of the designated dependence type to a
semantic role of a meta-task. For example, according to the first
rule in Table~\ref{tab:rules}, \textit{beverage} is assigned to the
semantic role \textit{Theme} of the meta-task \textit{put-Placing}.
Similarly, \textit{fridge} is assigned to \textit{Goal} of the same
meta-task according to the second rule. After matching, the single
instruction ``put beverage in the fridge'' is interpreted as an
instantiated meta-task of \textit{put-Placing} as follow:

\begin{small}
\begin{align*}
&(\ \textbf{define } (\ \textbf{meta-task } \textrm{put-Placing}\\
&\hspace{.5cm} (\ \textbf{:parameters } \texttt{robot beverage null fridge})) \\
&\hspace{.5cm} \textbf{... } \ )
\end{align*}
\end{small}

\begin{table}[t]
\begin{center}
\caption{Part of hierarchy for \textit{take-taking}.}\label{tab:role}
\centering
\begin{tabular}{ll}
\hline
\textbf{Semantic Role} & \textbf{Class} \\
\hline
\textit{Theme} & Holdable\_Obj \\
\textit{Source} & Supportable\_Obj $\sqcup$ Containable\_Obj \\
\hline
\end{tabular}\vspace{-10pt}
\end{center}
\end{table}

In the case where a semantic role of a sentence cannot be
identified within the sentence, semantic matching is conducted
based on a taxonomical hierarchy, which specifies what sorts of
entities can be taken as values by a semantic role. For example,
the \textit{Theme} role of meta-task \textit{put-Placing} should
take a holdable object for the robot. Table~\ref{tab:role} shows a
part of the hierarchy about meta-task \textit{take-Taking}.
Moreover, the hierarchy needs to be extended by class-subclass
relationships, as exemplified in Table~\ref{tab:tax}. Consider the
example sentence ``take the \textit{beer}'' in Figure 2. The
entities appeared in the context are \textit{fridge} and
\textit{fridge-door}. In our taxonomical hierarchy,
\textit{fridge-door} is an instance of \textit{door} which is
neither supportable nor containable. Therefore, only
\textit{fridge} can be a value of the \textit{Source} role of
\textit{take-Taking}. In the case of multiple candidates for a
semantic role, the nearest entity will be selected. The high-level
part of our hierarchy is similar to that of
AfNet~\cite{Varadarajan2012}. This is beneficial to integrating
grounding mechanism into our prototype system.

\begin{table}[t]
\begin{center}
\caption{Part of hierarchy for classes.}\label{tab:tax}
\centering
\begin{tabular*}{0.75\columnwidth}{@{\extracolsep{\fill}}lll}
\hline
\textbf{Class} & \textbf{Subclass} & \textbf{Subsubclass} \\
\hline
Object & Containable\_Obj & \textit{fridge} \\
Object & Holdable\_Obj & \textit{beer,beverage}\\
Object & Supportable\_Obj &  \textit{table} \\
\hline
\end{tabular*}\vspace{-10pt}
\end{center}
\end{table}

\section{Task Planning with ASP}
\label{sec:planning}

\noindent Given the meta-task semantic representation of a
sentence, we generate an action sequence using OMICS and functional
definition knowledge of common verbs (e.g., {\em Re-FrameNet}). In
our previous work, we proposed the {\em OK-planner}~\cite{Chen2012}
based on ASP. In this approach, all types of knowledge are
converted into ASP and then an ASP solver is applied to generate an
action sequence. However, this work does not consider {\em common
verbs} for handling complex tasks.

In this article, we built our planner upon our previous work but
additionally consider the following challenges: 1) how to define
the functional knowledge of primitive actions in \textit{Action
Model} and 2) how to convert \textit{Re-FrameNet} definition of
common verbs into ASP.

\subsection{Planning with Action Model}

\noindent As aforementioned, we specify robot skills in our system
by an action model, i.e., a set of primitive actions that are
executable for the robot. Table~\ref{tab:am} shown some basic
definition of the primitive actions for a typical service robot
though different types of robots may have different action model.
Formally, each primitive action $a$ is defined as a pair $\langle
pre(a),eff(a) \rangle$, where $pre(a)$ and $eff(a)$ are the
preconditions and effects of $a$ respectively. For instance,
\texttt{moveto(obj)} is a primitive action that tells the robot to
move close to the specified object \texttt{obj}. The $pre$ and
$eff$ of \texttt{moveto(obj)} show whether the robot is near the
specified \texttt{obj} before and after the \texttt{moveto} action
respectively.

Given any initial state $s_0$ and a possible plan $a_1$, $\ldots$,
$a_n$, an action model determines a predicted trajectory $\tau^* =
\langle s_0, a_1, s_1, \ldots, a_n, s_n\rangle$ through inference
for all the states $s_1$, $\ldots$, $s_n$ along with the execution
of the action sequence during planning. For instance, given an
instruction ``get food from fridge'', we need to generate a plan
for the robot as:

\begin{center}\small\tt
moveto(fridge,1), open(fridge,2), find(food,3), pick\_up(food,4),
close(fridge,5).
\end{center}

Note that the semantic representation of a user instruction can be
easily converted into a ASP form~\cite{Chen2012}. All we have to do
is to fill sufficient knowledge for the ASP planner. Using our
\textit{Re-FrameNet} definition, an action verb is reorganized by a
set of precondition, postcondition, and invariant over semantic
roles of the action. Therefore, the remaining problem for our
approach is how to convert the functional definitions of common
verbs into ASP.

\begin{table}
\begin{center}
\caption{List of primitive actions that can be executed by the robot.}\label{tab:am}\vspace{-10pt}
\centering
\begin{tabular}{|l|l|}
\hline
{Primitive Action(a)} & {Description(a), $pre(a), eff(a)$} \\
\hline
moveto(obj, t) & Move to obj by using motion planner at time t.\\
& $pre(a): not~near(robot, obj, t-1)$\\
& $eff(a): near(robot, obj, t)$\\ \hline
find(obj, t) & Find obj in the environment by using vision at time t.\\
& $pre(a): near(robot, obj, t-1)$\\
& $eff(a): beliveloction(robot, obj, t)$\\ \hline
pick\_up(obj, t) & pick up obj by using robotic arm at time t.\\
& $pre(a): near(robot, obj, t-1)$\\
& $pre(a): beliveloction(robot, obj, t-1)$\\
& $eff(a): grapsing(robot, obj, t)$\\ \hline
put\_down(obj, t) & put down obj on a plane in front of robot at time t.\\
& $pre(a): grapsing(robot, obj, t-1)$\\
& $eff(a): not ~grasping(robot, obj, t)$\\ \hline
open(obj, t) & open the obj at time t.\\
& $pre(a): closed(obj, t-1)$\\
& $eff(a): opened(obj, t)$\\ \hline
close(obj, t) & close the obj at time t.\\
& $pre(a): opened(obj, t-1)$\\
& $eff(a): closed(obj, t)$\\
\hline
\end{tabular}\vspace{-10pt}
\end{center}
\end{table}

\subsection{Conversion of Functional Knowledge}

\noindent Let $\alpha$ be a common verb (word sense). The set of
linguistic variables of $\alpha$'s frame is denoted by
$\Theta(\alpha)$. The set of properties and relations over
$\Theta(\alpha)$ occur in the functional definitions of verbs
belonging to $\alpha$'s Frame is denoted by $\Sigma(\alpha)$. Given
a task $\taska$ based on the common verb~$\alpha$ as:
\[\small
( \textbf{:meta-task } \alpha\ (\textbf{:parameters } (p_1\ \mathcal{X})\ \cdots (p_h\ \mathcal{X})))
\]
where $X_1, \ldots, X_h \in \Theta(\alpha)$ and $p_1, \ldots, p_h$
are predicates over a set $\mathcal{X}$ of variables, each
constraint of the common verb $\alpha$ can be converted to a set of
ASP rules w.r.t. the task $\taska$ as:

1. A precondition
\[\small
\hspace{-2em}(\textbf{:precond}\ \alpha\ (\textbf{conj } (\textbf{disj } l_1\ \cdots\ l_n) \cdots (\textbf{disj } l'_1\ \cdots\ l'_m)))
\]
is converted to the following ASP rules:
\begin{small}
\begin{align*}
\hspace{-3em}  \gets&\ \process(\taska, t, t'),\,\Not \true(l_1, t),\, \ldots,\, \Not\true(l_n, t),\\&\ t<t',\, p_1(\mathcal{X}),\, \ldots,\, p_h(\mathcal{X})\\
    &\cdots\\
\hspace{-3em}\gets&\ \process(\taska, t, t'),\, \Not\true(l'_1, t),\, \ldots,\, \Not\true(l'_n, t),\\&\ t<t',\, p_1(\mathcal{X}),\, \ldots,\, p_h(\mathcal{X})
\end{align*}
\end{small}

2. A postcondition
\[\small
\hspace{-2em} (\textbf{:postcond}\ \alpha\ (\textbf{conj } (\textbf{disj } l_1\ \cdots\ l_n) \cdots (\textbf{disj } l'_1\ \cdots\ l'_m)))
\]
is converted to the following ASP rules:
\begin{small}
\begin{align*}
\hspace{-3em} \gets&\ \process(\taska, t, t'),\,\Not\true(l_1, t'),\, \ldots,\, \Not\true(l_n, t'),\\&\ t<t',\, p_1(\mathcal{X}),\, \ldots,\, p_h(\mathcal{X})\\
    &\cdots\\
\hspace{-3em}  \gets&\ \process(\taska, t, t'),\, \Not\true(l'_1, t'),\, \ldots,\, \Not\true(l'_n, t'),\\&\ t<t',\, p_1(\mathcal{X}),\, \ldots,\, p_h(\mathcal{X})
\end{align*}
\end{small}

3. An invariant
\[\small
\hspace{-2em} (\textbf{:invariant}\ \alpha\ (\textbf{conj } (\textbf{disj } l_1\ \cdots\ l_n) \cdots (\textbf{disj } l'_1\ \cdots\ l'_m)))
\]
is converted to the following ASP rules:
\begin{small}
\begin{align*}
\hspace{-3em} \gets&\ \process(\taska, t, t'),\,\Not\true(l_1, t''),\, \ldots,\, \Not\true(l_n, t''),\\&\ t<t',\, t <= t'',\, t'' <= t',\, p_1(\mathcal{X}),\, \ldots,\, p_h(\mathcal{X})\\
    &\cdots\\
\hspace{-3em} \gets&\ \process(\taska, t, t'),\, \Not\true(l'_1, t''),\, \ldots,\, \Not\true(l'_n, t''),\\&\ t<t',\, t<=t'', \, t''<=t', \, p_1(\mathcal{X}),\, \ldots,\, p_h(\mathcal{X})
\end{align*}
\end{small}

4. An invariant
\begin{small}
\begin{multline*}
\hspace{-1em} ( \textbf{disj }(\textbf{:invariant}\ \alpha\ (\textbf{conj } (\textbf{disj } l_1\ \cdots\ l_n) \cdots (\textbf{disj } l'_1\ \cdots\ l'_m)))\\ \hspace{-2em} (\textbf{:invariant}\ \alpha\ (\textbf{conj } (\textbf{disj } l^*_1\ \cdots\ l^*_n) \cdots (\textbf{disj } l^{\prime*}_1\ \cdots\ l^{\prime*}_m))))
\end{multline*}
\end{small}
is converted to the following ASP rules:
\begin{small}
\begin{align*}
f\gets&\ \process(\taska, t, t'),\,\Not\true(l_1, t''),\, \ldots,\, \Not\true(l_n, t''),\\&\ t<t',\, t <= t'',\, t'' <= t',\, p_1(\mathcal{X}),\, \ldots,\, p_h(\mathcal{X})\\
    &\cdots\\
f\gets&\ \process(\taska, t, t'),\, \Not\true(l'_1, t''),\, \ldots,\, \Not\true(l'_n, t''),\\&\ t<t',\, t<=t'', \, t''<=t', \, p_1(\mathcal{X}),\, \ldots,\, p_h(\mathcal{X})\\
f^*\gets&\ \process(\taska, t, t'),\,\Not\true(l^*_1, t''),\, \ldots,\, \Not\true(l^*_n, t''),\\&\ t<t',\, t <= t'',\, t'' <= t',\, p_1(\mathcal{X}),\, \ldots,\, p_h(\mathcal{X})\\
    &\cdots\\
f^*\gets&\ \process(\taska, t, t'),\, \Not\true(l^{\prime*}_1, t''),\, \ldots,\, \Not\true(l^{\prime*}_n, t''),\\&\ t<t',\, t<=t'', \, t''<=t', \, p_1(\mathcal{X}),\, \ldots,\, p_h(\mathcal{X})\\
    \gets&\ f, f^*
\end{align*}
\end{small}

After all pieces of knowledge have been converted into the ASP
rules, an ASP solver \textit{iclingo}~\cite{gebser2008engineering}
--- a combination of \textit{Gringo} and \textit{clasp} for
incremental grounding and solving --- is used to incrementally
ground the ASP rules above and search for answer sets, from which a
plan can be computed~\cite{Chen2012}.

\section{Experiments}
\label{sec:experiments}

\noindent We empirically evaluate our system with three
experiments. The first experiment was devised to investigate the
performance of our SMR (i.e., Semantic Matching and Recovering)
method. The second experiment aimed to testing the performance of
the whole system when different open knowledge bases were used. We
also analyzed the main factors that may affect the performance.
Finally, we demonstrate that how our approach can be deployed in
our KeJia robot to solve instruction understanding problems in two
domestic scenarios. Additionally, we also present our long-term
effort on applying the proposed technique in the RoboCup@Home
competitions.

\subsection{Experiments with SMR}

\noindent To test our SMR method, we collect 191,740 examples
annotated with frame-semantic structures for the frame
identification model from {\em FrameNet} lexicon and 470 examples
from OMICS. Then, we parse each sentence by the Stanford parser.
Finally, we only select those examples whose LU is a verb or a verb
phrase. As a result, the training data contains 70,149 examples and
the test data contains 18,183 examples from {\em FrameNet} and 630
examples from OMICS. In our experiments, the frame identification
model instantiates 76,289 binary features.

Table~\ref{tab:smr} shows the results on each part of translation
of hierarchal instructions. The performance is evaluated by {\em
Precise} (P), {\em Recall} (R), {\em F1} (F) defined as: $Precise =
TP/(TP+FP)$, $Recall = TP/T$, $F1 =2*Precise*Recall /(Precise +
Recall)$, where $TP$ stands for the number of the sentences parsed
correctly, $FP$ is the number of the sentences parsed wrongly, and
$T$ is the total length of the dataset.

As we can see from the results, syntactic results have a very high
precise and $F1$ value, which benefits to the meta-task
identification phase. However, it does not disambiguate the meaning
of a verb (e.g., the verb ``get'' has two meanings: ``Getting: get
the food'' and ``Motion: get to the room''). The meta-task
identification, which obtains a $F1$ value of 80 over the {\em
FrameNet} data and 71.07 over the OMICS data. Moreover, the overall
performance of the whole translation system maintains a quite high
precise and relatively low recall due to the data sparseness and
one meta-task assumption.

\begin{table}
\begin{center}
\caption{Results of translation over two testsets of FrameNet and OMICS.}\label{tab:smr}
\centering
\begin{tabular*}{0.9\columnwidth}{@{\extracolsep{\fill}}|l|l|l|l|l|}
\hline
\textbf{Syntactic} & \textbf{Data} & \textbf{P} & \textbf{R} & \textbf{F} \\
\hline
Verb & OMICS & 97.61 & 81.83 & 89.03\\
Entities & OMICS & 80.32 & 67.33 & 73.25\\
\hline
\textbf{Identification} & \textbf{Data} & \textbf{P} & \textbf{R} & \textbf{F} \\
\hline
Frame & OMICS & {84.31} & {61.43} & {71.07} \\
Frame & FrameNet & {80.98} & {79.05} & {80.00} \\
Semantic Roles & OMICS & {78.00} & {53.71} & {63.62} \\
\hline
\end{tabular*}\vspace{-10pt}
\end{center}
\end{table}

\subsection{Experiments on OMICS}

\noindent The experiments on OMICS were divided into two tests.
Test 1 was conducted on 11,885 user tasks from the
\textit{Tasks/Steps} table and Test 2 on 467 user desires from the
\textit{Help} table.

Test 1 consisted of four rounds. In the first round, only the
definitions of the 11,885 tasks from the \textit{Tasks/Steps} table
and a small action model \textit{AM} representing the basic
perception and manipulation skills of a robot were used.
Specifically, \textit{AM} contained only 6 primitive actions:
\textit{move, find, pick\_up, put\_down, open}, and \textit{close}.
Synonymy knowledge from \textit{FrameNet} was used into the second
to fourth rounds of Test 1. In the third and fourth rounds,
rewritten knowledge from \textit{Re-FrameNet} was considered with
our SMR technique. However, in the third round, missing roles were
not recovered from the context.

\begin{table}
\begin{center}
\caption{Experimental results over 11885 user tasks.}\label{tab:Taskres}
\centering
{\footnotesize
\begin{tabular*}{0.9\columnwidth}{@{\extracolsep{\fill}}|l||c|c|c|c|}
\hline
\textbf{Test 1} & {AM} & {FN} & {SMR\_0} & \textbf{SMR\_1} \\
\hline
\hline
\textit{Tasksteps} & {134} & {150} & {618} & \textbf{790} \\
\hline
\textit{Tasksteps+} & {157} & {174} & {756} & \textbf{935} \\
\hline
\textit{Percent(\%)} & {1.32} & {1.46} & {6.36} & \textbf{7.87} \\
\hline
\textit{GroundTruth(\%)} & {*} & {*} & {63.75} & {64} \\
\hline
\textit{TruthPercent(\%)} & {1.32} & {1.46} & {4.05} & \textbf{5.04} \\
\hline
\end{tabular*}}
\end{center}
\end{table}

Table~\ref{tab:Taskres} shows the experimental results of Test 1.
The second row shows the numbers of tasks that were successfully
planned by the \textit{global} planner with tasks/steps in the four
rounds. The third row shows the total numbers of tasks that were
successfully planned in the four rounds. The fourth row shows the
percentages of successfully planned tasks with respect to the total
number of tested tasks. Since there are no ground truth data for
OMICS, we randomly drew 80 and 100 samples from the last two rounds
respectively and verified them manually. It turned out that 51 and
64 samples among them were correct. As shown in the fifth row of
Table~\ref{tab:Taskres}, the correctness percent decreased when
{\em Re-FrameNet} was used; but the number of correctly planned
tasks still increased remarkable. Moreover, we can see that the
overall performance improved when semantic roles of common verbs
was used, much better than the state-of-art
solution~\cite{Chen2012}.

\begin{figure}
\centering
\subfigure[]{
\label{fig:trend}
\includegraphics[width=0.36\columnwidth]{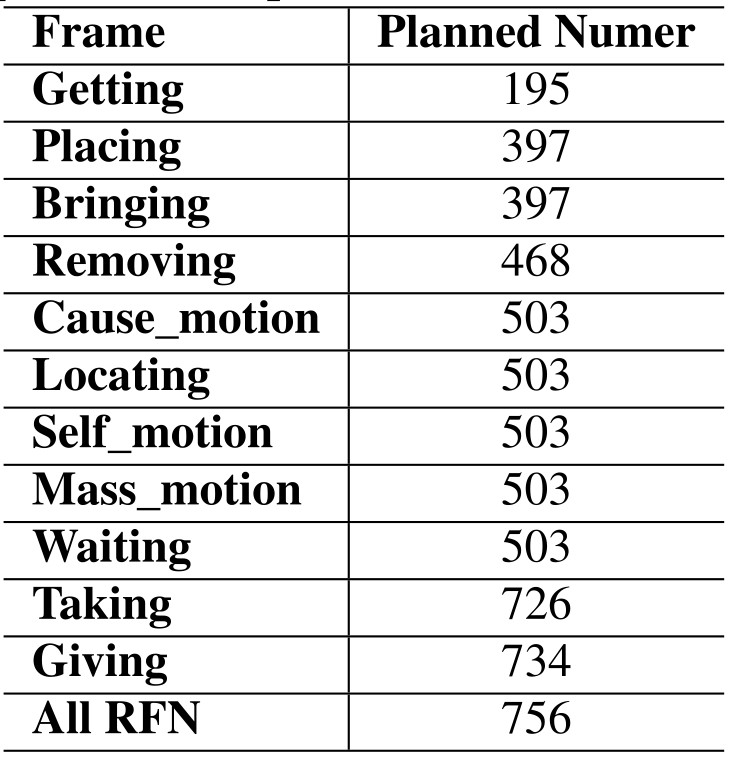}
}
\subfigure[]{
\label{fig:taskresults}
\includegraphics[width=0.57\columnwidth]{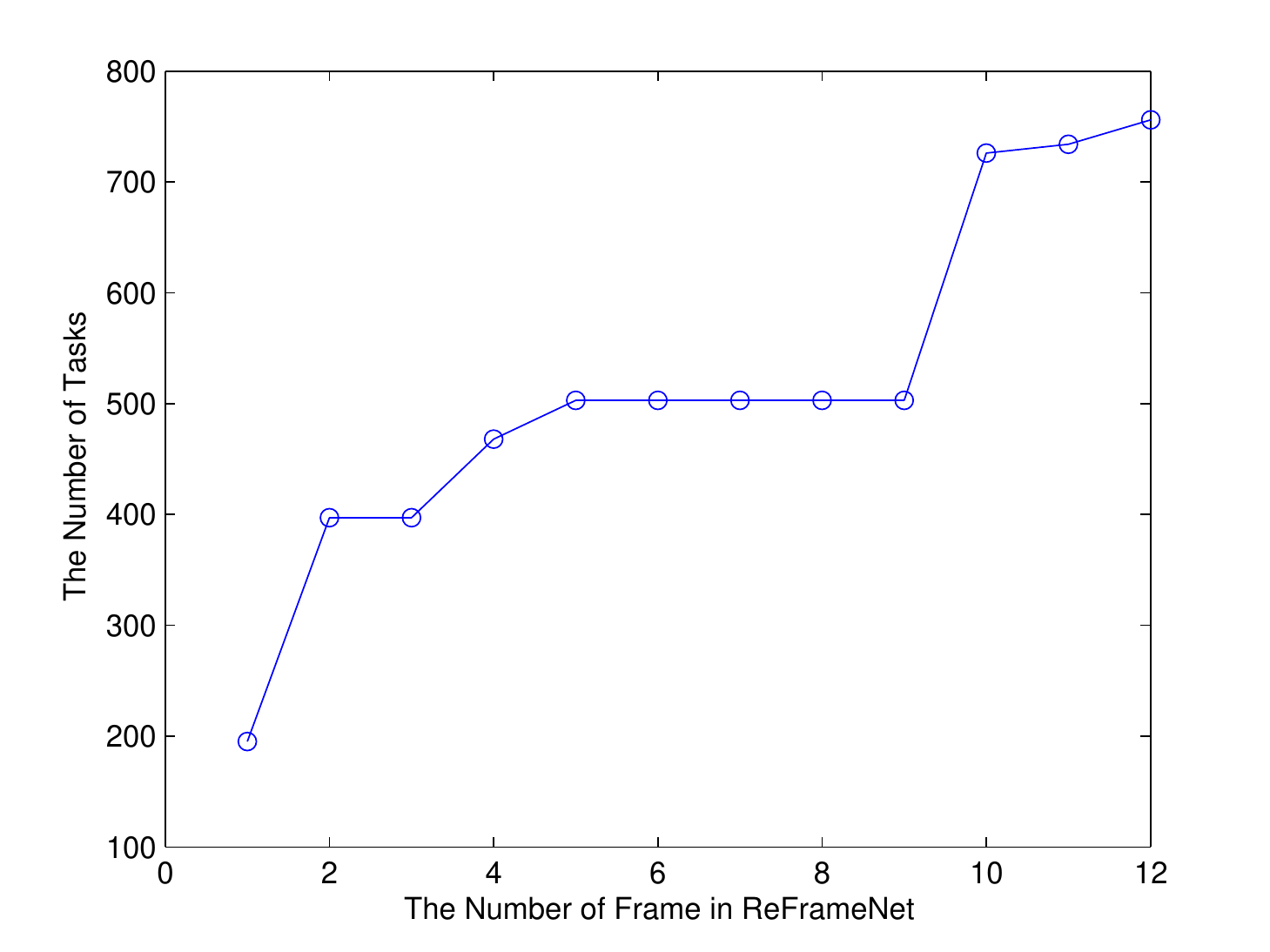}
}
\caption{Influences of the Frame in Re-FrameNet in Test 1.}
\label{fig:overall}\vspace{-10pt}
\end{figure}

As shown in Figure~\ref{fig:overall}, the number of the
successfully planned tasks gradually increased when more frames
were added to the algorithm. It also shows that some frames cannot
be mapped into robots' action (i.e., \textit{Mass\_motion} and
\textit{Waiting}). The main reason is the limit of robots'
primitive actions.

Table~\ref{tab:Failres} reports the main types of failures that we
observed in Test 1. Specifically, the {\em Parsed Failure} occurred
in 3027 tasks because the semantic matching and recovering
procedure failed to retrieve any frame from \textit{Re-FrameNet}
(RFN) for a task. The {\em RFN Failure} occurred in 4394 tasks due
to the fact that {\em Re-FrameNet} contains only 43 frames, in
which 7421 tasks cannot be used to generate a plan by the robot. A
{\em Global Planning Failure} occurs when a task/step \textit{t}
cannot be planned and none of the following conditions hold:
\textit{t} is a primitive action, semantically equivalent to
meta-task in \textit{Re-FrameNet} or another task in the {\em
Tasks/Steps} table. In total, there were 3527 tasks failed in this
category. A {\em Local Planning Failure} occurs when the $solver$
(in Algorithm 3) is launched but fails to generate any plan.
Further study reveals that these two sorts of planning failures are
mainly due to lack of knowledge/skills.

Test 2 was conducted on 467 user desires from the \textit{Help}
table of OMICS. The experimental results are shown in
Table~\ref{tab:Desres}. As we can see, the success rates were
higher than the corresponding rounds of Test 1. In particular, the
success rate is as high as 81\% in the last round. This is because
a desire can be met by various tasks, which can be different from
one another. Therefore, knowledge used in the rounds of Test 2 was
much richer than that in Test 1.

\begin{table}
\begin{center}
\caption{Experimental results over 467 user desires.}\label{tab:Desres}
\centering
{\footnotesize
\begin{tabular*}{0.9\columnwidth}{@{\extracolsep{\fill}}|l||c|c|c|c|}
\hline
\textbf{Test 2} & {AM} & {FN} & {SMR\_0} & \textbf{SMR\_1} \\
\hline
\hline
\textit{Help} & {244} & {247} & {299} & \textbf{331} \\
\hline
\textit{Help+Tasksteps} & {254} & {261} & {358} & \textbf{379} \\
\hline
\textit{Percent(\%)} & {54.39} & {55.89} & {76.66} & \textbf{81.16} \\
\hline
\end{tabular*}}
\end{center}
\end{table}

\begin{table}
\begin{center}
  \caption{Influences of main factors of failure in Test 1.}\label{tab:Failres}
\begin{tabular*}{0.9\columnwidth}{@{\extracolsep{\fill}}l|c|c}
{\bf Failure} & {\bf Number} & {\bf Percent (\%)}  \\
\hline
Parsed Failure & 3027 & 26.7 \\
RFN Failure & 4394 & 38.8 \\
Global Planning Failure & 3527 & 31.2 \\
Local Planning Failure & 378 & 3.3 \\
\end{tabular*}\vspace{-10pt}
\end{center}
\end{table}

Notice that the overall performance increased about 5 times in Test
1 and 50\% in Test 2 when semantic roles of common verbs and
\textit{Re-FrameNet} was used. There are two main reasons for this
improvement. Firstly, rewritten knowledge of common verbs in
\textit{Re-FrameNet} fills knowledge gaps caused by lack of
definitions of these verbs in OMICS . Without the knowledge,
761($=$935-174) tasks would not have been successfully planned in
the last two rounds of Test 1. Secondly, our SMR mechanism
contributed significantly to the improvement. Without it,
179($=$935-756) out of these 761 tasks would not have been
successfully planned. In other words, \textit{Re-FrameNet} and SMR
made about 76\% and 24\% contributions to the improvement of
success rate in task planning respectively.

\subsection{Case Study on KeJia Robot}

\begin{figure}[t]
\centering
\subfigure[(move(loc(floor)),1)]{
\label{fig:clean_pick_move}
\includegraphics[width=0.4\columnwidth]{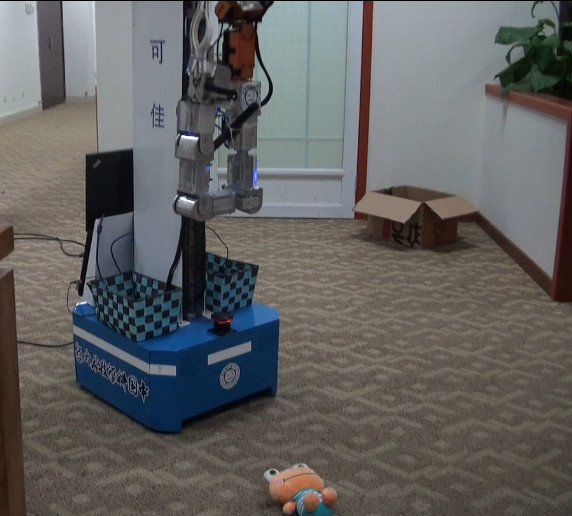}
}
\subfigure[(pick\_up(toy),2)]{
\label{fig:clean_pick_up}
\includegraphics[width=0.4\columnwidth]{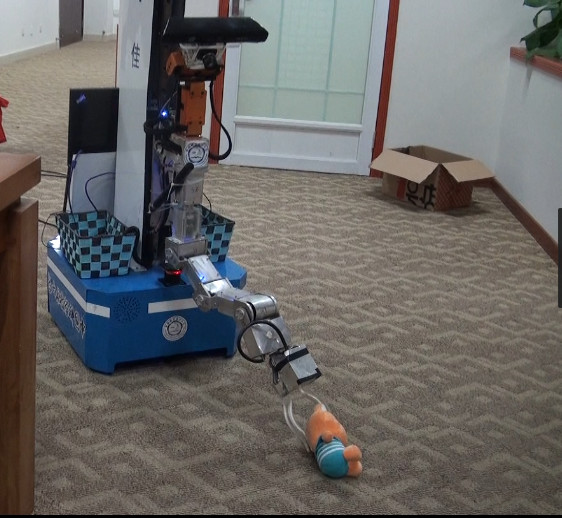}
}
\subfigure[(move(loc(toybox)),3)]{
\label{fig:clean_put_move}
\includegraphics[width=0.4\columnwidth]{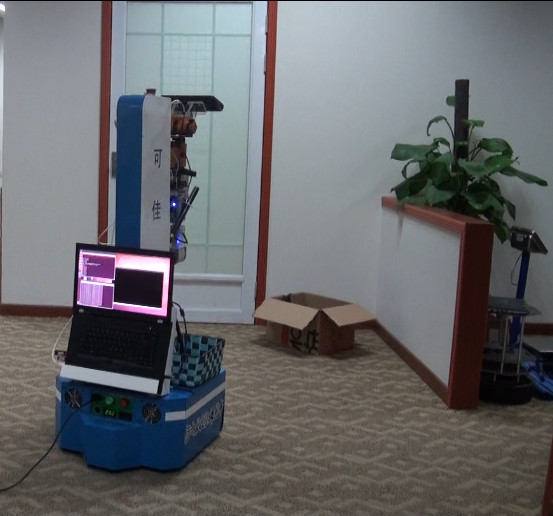}
}
\subfigure[(put\_down(toy),4)]{
\label{fig:clean_put_down}
\includegraphics[width=0.4\columnwidth]{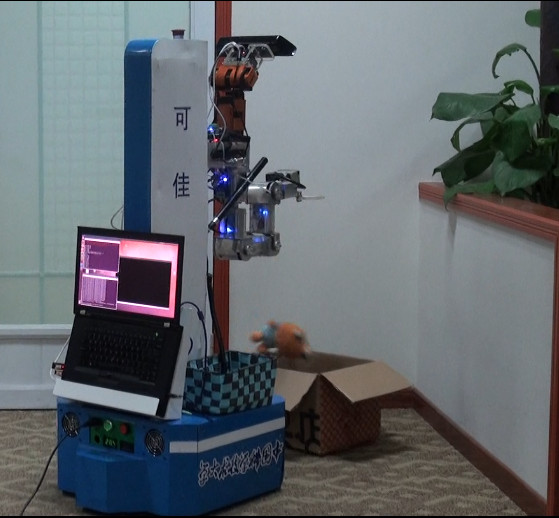}
}
\caption{Execution of the task ``clean up toys'' in tasksteps.
subfigure (a) and (b) are plans for ``pick up toys from floor'',
(c) and (d) for ``put toys in toybox''.}
\label{fig:taskstep}\vspace{-10pt}
\end{figure}

\begin{figure}[t]
\centering
\subfigure[(move(loc(aspirin)),1)]{
\label{fig:pick_move}
\includegraphics[width=0.4\columnwidth]{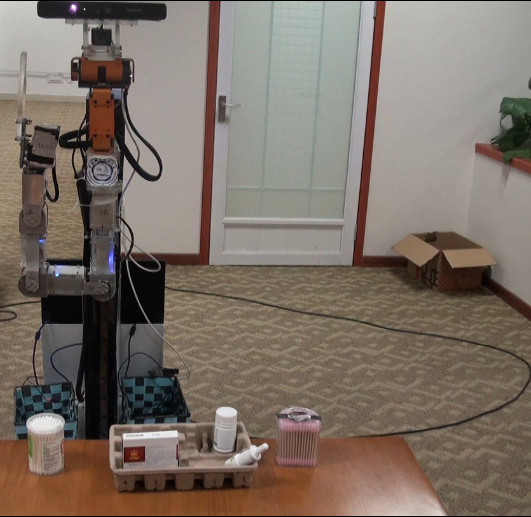}
}
\subfigure[(pick\_up(aspirin),2)]{
\label{fig:pick_up}
\includegraphics[width=0.4\columnwidth]{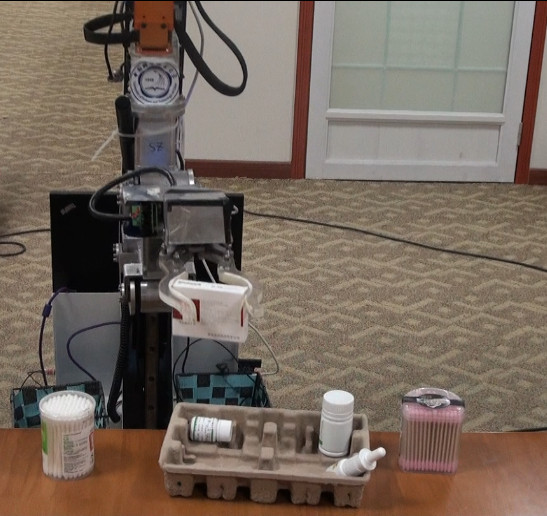}
}
\subfigure[(move(loc(them)),3)]{
\label{fig:put_move}
\includegraphics[width=0.4\columnwidth]{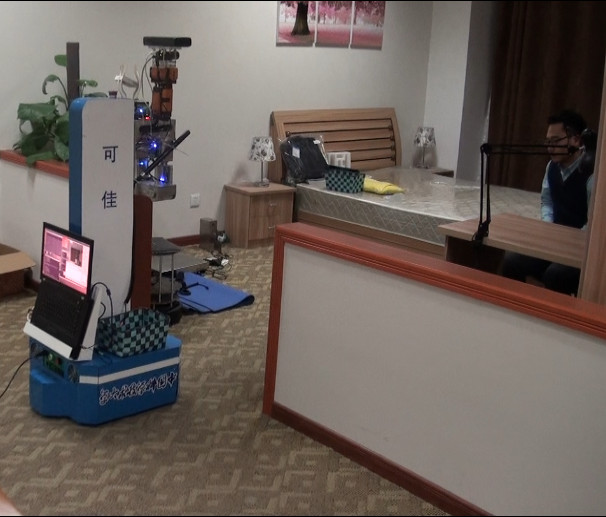}
}
\subfigure[(put\_down(aspirin),4)]{
\label{fig:put_down}
\includegraphics[width=0.4\columnwidth]{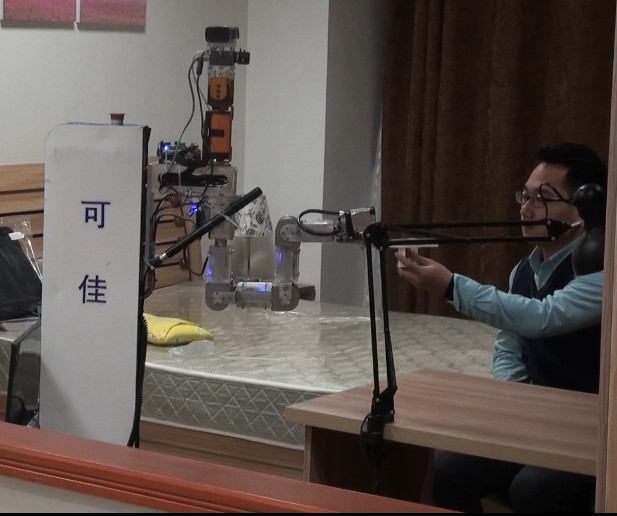}
}
\caption{Execution of ``give them an aspirin'' for the desire ``have a headache''.}
\label{fig:help}
\end{figure}

\noindent We conducted a case study of our system with the KeJia
robot. As shown in Figures \ref{fig:taskstep} and \ref{fig:help},
our KeJia robot is based on a two-wheels driving chassis of
62cm$\times$53cm$\times$32cm and the equipped sensors include a
laser range finder, a 1394 camera, and a Kinect. A lifting system
is mounted on the chassis attached with the robot's upper body.
Assembled with the upper body is a 6 DOF arm. It is able to reach
objects over 83cm far from mounting point and the maximum payload
is about 500g. The robot's power is supplied by a 20Ah battery that
guarantees the continuous running of at least 1 hour. The
computational resources consist of a laptop and an on-board PC. Our
system is built upon existing modules including motion control for
the mobile base and arm, navigation, recognition and localization.

In our case study, we first tried two typical scenarios where the
robot can benefit from the proposed techniques. Then, we introduce
our long-term effort on developing general-purpose systems for user
instruction understanding in the annual RoboCup@Home competitions.

\subsubsection{Scenario 1}

As shown in Figure~\ref{fig:taskstep}, a toy and a toy box were
placed on the floor. Our KeJia robot was asked by a user to ``clean
up toys''. Note that, with only this instruction, the robot is
unable to complete the task because the action ``clean up'' is
unspecified. In our system, the robot first extracted the subtasks
of the task ``clean up toys" based on the knowledge in OMICS. By
doing so, a tuple of $\langle$\textit{task}. ``clean up toys'':
\textit{step 1}. ``pick up toys from floor''; \textit{step 2}.
``put toys in toybox''. $\rangle$ was generated. Then, our SMR
method matched and recovered semantic roles of each step in the
tuple as:
\begin{small}
\begin{align*}
&(\ \textbf{define } (\ \textbf{task } \textrm{clean\_up (toys)}\\
&\hspace{.5cm} (\ \textbf{:subtasks } \textrm{pick\_up-Pick\_up} \\
&\hspace{.5cm} \hspace{.5cm} (\ \textbf{:parameters } \texttt{toys floor} ) ) \\
&\hspace{.5cm} (\ \textbf{:subtasks } \textrm{put-Placing} \\
&\hspace{.5cm} \hspace{.5cm} (\ \textbf{:parameters } \texttt{toys floor toybox} ) )
) )
\end{align*}
\end{small}

After that, our planner sequentially processed each subtask. In
this phase, since the action {\em pick\_up} is a primitive action,
the subtask {\em pick\_up} can be directly executed by our robot.
For the second subtask, we tried to generate a plan given the
definition of the meta-task {\em put-Placing} as:
\begin{small}
\begin{align*}
&(\ \textbf{define } (\ \textbf{meta-task } \textrm{put-Placing}\\
&\hspace{.5cm} (\ \textbf{:parameters } \textit{?Agent ?Theme ?Source ?Goal}))\\
&\hspace{.5cm} (\ \textbf{:precondition  } \textit{(at Theme Source)} )\\
&\hspace{.5cm} (\ \textbf{:precondition  } \textit{(conj(portable Theme)(object Theme))} )\\
&\hspace{.5cm} (\ \textbf{:postcondition  } \textit{(at Theme Goal)} )
\end{align*}
\end{small}

In this scenario, the plan generated by the planner for this task
is shown in Figures~\ref{fig:clean_put_move} and
\ref{fig:clean_put_down}. At this point, the task ``clean up toys''
is solved by our system and finally the entire plan is executed by
the robot to complete the task.

\subsubsection{Scenario 2}

As shown in Figure~\ref{fig:help}, a user spoke to the robot that
he ``have a headache''. This was identified as a user desire.
Similar to the previous scenario, our system first extracted a
series of help tasks for the user desire such as ``with pain
medication'', ``give them an aspirin'', etc. Then, our SMR method
matched and recovered semantic roles of each help task. In this
scenario, our planner failed to plan for the task ``with pain
medication'' but successfully recovered the \textit{Source}
elements and generated a plan for the task ``give them an
aspirin''. A list of actions for the plan of this task are
illustrated in Figure~\ref{fig:help}.

A video demon for the two scenarios above with our KeJia robot is
given at: {\tt https://youtu.be/A4GBXHG0l74}

\subsubsection{RoboCup@Home}

This is an international annual competition for domestic service
robots and is part of the RoboCup event. In this competition, a set
of benchmark tests are proposed to evaluate the robots' abilities
and performance in a realistic non-standardized home environment
setting. The most related benchmark test to this article is the
{\em General Purpose Service Robot} (GPSR) test, which requires a
robot to solve tasks upon request in natural language randomly
generated by the referees during the competition.

In the RoboCup@Home competitions of the past three years, our team
--- WrightEagle (WE)~\cite{BWCrobocup12} got the 1st place once and 2nd place twice.
Table~\ref{tab:robocup} shows the total scores of the top 5 teams
in the benchmark tests (without the final stage). It can be seen
from the results that our team (i.e., WE) performed very well in
the competitions. Particularly, in the GPSR tests, the performance
of our system was competitive comparing to other top teams as shown
in Table~\ref{tab:gpsr}.

\begin{table}
\begin{center}
\caption{Scores of All RoboCup@Home Benchmark
Tests.}\label{tab:robocup} \centering
\begin{tabular}{|l|l|l|l|l|l|}
\hline
\textbf{Competition } & {top 1} & {top 2} & {top 3} & {top 4} & {top 5} \\
\hline
\hline
{RoboCup 2013} & {4767} & {4645} & {3622} & {3155} & {3066} \\
Team Name & \textbf{WE} & NimbRo & TU/e & Homer & BORG \\
\hline
{RoboCup 2014} & {9305} & {5701} & {5656} & {4842} & {3417} \\
Team Name & \textbf{WE} & TU/e & NimbRo & Tobi & Pumas \\
\hline
{RoboCup 2015} & {750} & {651} & {647} & {562} & {359} \\
Team Name & \textbf{WE} & Homer & TU/e & Tobi & Pumas \\
\hline
\end{tabular}
\end{center}
\end{table}

Although there are generally many factors contributing to the
success in the RoboCup@Home competitions, our robot did benefit
substantially from the proposed system as described in this article
to process user instructions and generate plans. The competitions
motivated us to develop a general-purpose system for understanding
user instructions in natural language and also provide a good
testbed for such systems.

\section{Related Work}
\label{sec:related}

\noindent To date, many approaches on instruction understanding and
task planning for service robots have been proposed in the
literature. For instance, several integrated systems
\cite{Dzifcak2009,Cantrell2010,Kress2009} for natural language
understanding have been introduced to enable robots to complete
tasks given instructions in natural language. However, they all
assume that instructions are definitely specified for the domains
and do not consider semantic disambiguation of verbs and their
roles. Work have been proposed to manually create
environment-driven instructions for grounding user instructions in
natural language to robots' actions~\cite{Misra14,Hemachandra14}.
However, these methods cannot scale to large number of tasks
because each task need to be manually specified in an environment,
and are not suitable for different types of robots (e.g., robots
with different arm configurations).

To improve generality and scalability, researchers have tried to
exploit online knowledge and learn large-scare knowledge
representations to build a general-purpose system for instruction
understanding. For example, Lemaignan {\it et
al.}~\cite{Lemaignan2012,Lemaignanet2012} have tried to understand
and reason about knowledge around an action model using online
knowledge for robots. It is worth pointing out that we previously
proposed an integrated system~\cite{Chen2012} for our KeJia robot
consisting of multi-mode NLP, integrated decision-making, and open
knowledge searching.

The approaches that are most related to ours are the ones using
OMICS for robots to complete household tasks. The first attempt to
utilize OMICS to accomplish a household task is
\cite{shah2005building}, which proposed a generative model based on
the Markov chain techniques. Later on,
\cite{tenorth2010knowrob,kunze2010putting,tenorth2013knowrob}
presented a system called KNOWROB for processing knowledge in order
to achieve more flexible and general behavior. Most recently, we
proposed a formal description of knowledge gaps between user
instructions and local knowledge in robotic system for instruction
understanding
\cite{chen2012toward,Chen2012,xie2014understanding,xie2015multi}.
However, in these efforts using OMICS for robot task planning with
user instructions, {\em common verbs} are normally not defined in
the knowledge base, which limits their performance on utilizing
existing open knowledge. Thus, our work is proposed to address the
weakness of state-of-the-art methods.

\begin{table}
\begin{center}
\caption{Scores of The GPSR Benchmark Tests.}\label{tab:gpsr}
\centering
\begin{tabular}{|l|l|l|l|l|l|}
\hline
\textbf{GPSR Test } & {top 1} & {top 2} & {top 3} & {top 4} & {top 5} \\
\hline
\hline
{RoboCup 2013} & {900} & {500} & {450} & {250} & {250} \\
Team Name & NimbRo & Pumas & \textbf{WE} & TU/e & Tobi \\
\hline
{RoboCup 2014} & {750} & {700} & {500} & {0} & {0} \\
Team Name & \textbf{WE} & NimbRo & TU/e & Tobi & Pumas \\
\hline
{RoboCup 2015} & {105} & {60} & {30} & {30} & {20} \\
Team Name & Tobi & \textbf{WE} & TU/e & Homer & Skuba \\
\hline
\end{tabular}
\end{center}
\end{table}

\section{Conclusions}
\label{sec:conclusions}

\noindent This article proposed a general-purpose system for
service robot handling large-scale user instructions in natural
language. The key problem that we addressed is  how to map
primitive tasks into robot actions using semantic roles of common
verbs provided by semantic dictionaries --- a common resource of
open knowledge in linguistics. To solve this problem, we proposed a
novel approach for semantic matching and recovering. Furthermore,
we utilized semantic roles of common verbs defined in semantic
dictionaries for handling underspecification of naturalistic
language instructions in task planning. Empirical evaluation and
analysis were made and show good performance with two test sets
consisting of 11885 user tasks and 467 user desires collected from
OMICS. Moreover, we developed a prototype system deployed on our
KeJia robot and demonstrated our techniques with two typical
scenarios. Notably, our system has been used in the RoboCup@Home
competitions and shown good performance in the benchmark tests over
the past three years.

Here, we conclude with the following findings:
\begin{enumerate}
\item Overall performance of our system can be improved when
    \textit{Re-FrameNet} was used. As shown by our experimental
    results, both the knowledge in \textit{Re-FrameNet} and the
    SMR technique contributed to the improvement, indicating
    that rewritten knowledge of common verbs and recovering
    semantic roles from context are useful for naturalistic
    instruction understanding and planning.
\item The computational efficiency of our system can be
    improved using the hierarchism of user instructions and
    knowledge. As shown by our case study, instruction
    understanding and task planning can be done for our robot
    in realtime, given that task decomposition knowledge such
    as OMICS was used for efficient global planning and costly
    local planning was limited only to small number of
    low-level tasks defined in \textit{Re-FrameNet}.
\end{enumerate}

In the future, we plan to develop techniques to learn extra
knowledge unavailable from user input, such as knowledge about
robot manipulation, action configurations in finer degrees other
than semantic role, and most importantly grounding. Moreover, we
will investigate methods to automatically generate a large set of
{\em Re-FrameNet} for robot tasks.



\bibliographystyle{IEEEtran}
\bibliography{IEEE}

\begin{thebibliography}{10}
\providecommand{\url}[1]{#1}
\csname url@samestyle\endcsname
\providecommand{\newblock}{\relax}
\providecommand{\bibinfo}[2]{#2}
\providecommand{\BIBentrySTDinterwordspacing}{\spaceskip=0pt\relax}
\providecommand{\BIBentryALTinterwordstretchfactor}{4}
\providecommand{\BIBentryALTinterwordspacing}{\spaceskip=\fontdimen2\font plus
\BIBentryALTinterwordstretchfactor\fontdimen3\font minus
  \fontdimen4\font\relax}
\providecommand{\BIBforeignlanguage}[2]{{%
\expandafter\ifx\csname l@#1\endcsname\relax
\typeout{** WARNING: IEEEtran.bst: No hyphenation pattern has been}%
\typeout{** loaded for the language `#1'. Using the pattern for}%
\typeout{** the default language instead.}%
\else
\language=\csname l@#1\endcsname
\fi
#2}}
\providecommand{\BIBdecl}{\relax}
\BIBdecl

\bibitem{Chen2010}
X.~Chen, J.~Ji, J.~Jiang, G.~Jin, F.~Wang, and J.~Xie, ``{Developing High-level
  Cognitive Functions for Service Robots},'' in \emph{Proceedings of 9th
  International Conference on Autonomous Agents and Multi-agent Systems}, 2010.

\bibitem{Dzifcak2009}
J.~Dzifcak, M.~Scheutz, C.~Baral, and P.~Schermerhorn, ``{What to do and how to
  do it: Translating natural language directives into temporal and dynamic
  logic representation for goal management and action execution},'' in
  \emph{IEEE International Conference on Robotics and Automation}.\hskip 1em
  plus 0.5em minus 0.4em\relax ICRA, 2009, pp. 4163--4168.

\bibitem{Kollar2010}
T.~Kollar, S.~Tellex, D.~Roy, and N.~Roy, ``{Toward understanding natural
  language directions},'' in \emph{5th ACM/IEEE International Conference on
  Human-Robot Interaction}, 2010.

\bibitem{Nyga2012}
D.~Nyga and M.~Beetz, ``Everything robots always wanted to know about housework
  (but were afraid to ask),'' in \emph{IEEE/RSJ International Conference on
  Intelligent Robots and Systems}, 2012.

\bibitem{Saxena14}
A.~Saxena, A.~Jain, O.~Sener, A.~Jami, D.~K. Misra, and H.~S. Koppula,
  ``Robobrain: Large-scale knowledge engine for robots,'' in
  \emph{International Symposium of Robotics Research}, 2014.

\bibitem{Tellex2011}
S.~Tellex, T.~Kollar, S.~Dickerson, M.~Walter, A.~Banerjee, S.~Teller, and
  N.~Roy, ``{Understanding natural language commands for robotic navigation and
  mobile manipulation},'' in \emph{Proceedings of National Conference on
  Articial Intelligence}, 2011.

\bibitem{Gupta2004}
R.~Gupta and M.~Kochenderfer, ``{Common sense data acquisition for indoor
  mobile robots},'' in \emph{Proceedings of the 19th National Conference on
  Artificial Intelligence}, San Jose, California, USA, 2004, pp. 605--610.

\bibitem{Chen2012}
X.~Chen, J.~Ji, Z.~Sui, and J.~Xie, ``Handling open knowledge for service
  robots,'' in \emph{Proceedings of the Twenty-Third International Joint
  Conference on Artificial Intelligence}, 2013.

\bibitem{west1953general}
M.~P. West, \emph{A general service list of English words: with semantic
  frequencies and a supplementary word-list for the writing of popular science
  and technology}.\hskip 1em plus 0.5em minus 0.4em\relax Longmans, Green,
  1953.

\bibitem{Misra14}
D.~Misra, J.~Sung, K.~Lee, and A.~Saxena, ``Tell me dave: Context-sensitive
  grounding of natural language to manipulation instructions,'' in \emph{The
  International Journal of Robotics Research}, 2014.

\bibitem{Gelfond1988}
M.~Gelfond and V.~Lifschitz, ``The stable model semantics for logic
  programming,'' in \emph{Proceedings of the 5th International Conference on
  Logic Programming}.\hskip 1em plus 0.5em minus 0.4em\relax ICLP-88, 1988, pp.
  1070--1080.

\bibitem{Baker1998}
C.~F. Baker, C.~J. Fillmore, and J.~B. Lowe, ``{The berkeley framenet
  project},'' in \emph{Proceedings of the 17th international conference on
  Computational linguistics}.\hskip 1em plus 0.5em minus 0.4em\relax
  Association for Computational Linguistics, 1998, pp. 86--90.

\bibitem{bogaards1996dictionaries}
P.~Bogaards, ``Dictionaries for learners of english,'' \emph{International
  Journal of Lexicography}, vol.~9, no.~4, pp. 277--320, 1996.

\bibitem{DeMarneffe2006}
M.-C. de~Marneffe, B.~Maccartney, and C.~D. Manning, ``{Generating Typed
  Dependency Parses from Phrase Structure Parses},'' in \emph{Proceedings of
  the 5th International Conference on Language Resources and Evaluation
  (LREC-06)}.\hskip 1em plus 0.5em minus 0.4em\relax Genoa, Italy: ELRA/ELDA
  Paris, 2006, pp. 449--454.

\bibitem{DeMarneffe2008}
M.-C. de~Marneffe and C.~D. Manning, ``{The Stanford typed dependencies
  representation},'' in \emph{Proceedings of the COLING 2008 Workshop on
  Cross-framework and Cross-domain Parser Evaluation}, no.~ii.\hskip 1em plus
  0.5em minus 0.4em\relax Manchester, UK: ACL, 2008, pp. 1--8.

\bibitem{Cantrell2010}
R.~Cantrell, M.~Scheutz, P.~Schermerhorn, and X.~Wu, ``{Robust spoken
  instruction understanding for HRI},'' in \emph{Proceedings of the 5th
  ACM/IEEE International Conference on Robot Interaction}, 2010.

\bibitem{Kollar2013}
T.~Kollar, V.~Perera, D.~Nardi, and M.~Veloso, ``Learning environmental
  knowledge from task-based human-robot dialog,'' in \emph{Proc. of the IEEE
  International Conference on Robotics and Automation}, 2013.

\bibitem{Varadarajan2012}
K.~M. Varadarajan and M.~Vincze, ``{AfRob: The Affordance Network Ontology for
  Robots},'' in \emph{IEEE/RSJ International Conference on Intelligent Robots
  and Systems}, 2012.

\bibitem{Williams2013}
T.~Williams, R.~Cantrell, G.~Briggs, P.~Schermerhorn, and M.~Scheutz,
  ``{Grounding Natural Language References to Unvisited and Hypothetical
  Locations},'' in \emph{Proceedings of the Twenty-Seventh AAAI Conference on
  Artificial Intelligence}, Bellevue, Washington, USA, 2013.

\bibitem{gebser2008engineering}
M.~Gebser, R.~Kaminski, B.~Kaufmann, M.~Ostrowski, T.~Schaub, and S.~Thiele,
  ``Engineering an incremental asp solver,'' in \emph{Logic Programming}.\hskip
  1em plus 0.5em minus 0.4em\relax Springer, 2008, pp. 190--205.

\bibitem{BWCrobocup12}
A.~Bai, F.~Wu, and X.~Chen, ``Towards a principled solution to simulated robot
  soccer,'' in \emph{Proceedings of the Robot Soccer World Cup XVI Symposium
  (RoboCup)}, Mexico City, Mexico, 2012, pp. 141--153.

\bibitem{Kress2009}
H.~Kress-Gazit, G.~E. Fainekos, and G.~J. Pappas, ``Temporal-logic-based
  reactive mission and motion planning,'' \emph{IEEE Transactions on Robotics},
  vol.~25, no.~6, pp. 1370--1381, 2009.

\bibitem{Hemachandra14}
S.~Hemachandra, M.~Walter, S.~Tellex, and S.~Teller, ``Learning
  spatial-semantic representations from natural language descriptions and scene
  classifications,'' in \emph{2014 IEEE International Conference on Robotics
  and Automation (ICRA)}, 2014, pp. 2623--2630.

\bibitem{Lemaignan2012}
S.~Lemaignan, ``Grounding the interaction: knowledge management for interactive
  robots,'' \emph{KI-K unstliche Intelligenz}, pp. 1--3, 2012.

\bibitem{Lemaignanet2012}
S.~Lemaignan, R.~Ros, E.~Sisbot, R.~Alami, and M.~Beetz, ``Grounding the
  interaction: Anchoring situated discourse in everyday human-robot
  interaction,'' \emph{International Journal of Social Robotics}, vol.~4,
  no.~2, pp. 181--199, 2012.

\bibitem{shah2005building}
C.~Shah and R.~Gupta, ``Building plans for household tasks from distributed
  knowledge,'' in \emph{Proceedings of the 19th International Joint Conference
  on Artificial Intelligence (IJCAI 2005) Workshop on Modeling Natural Action
  Selection}.\hskip 1em plus 0.5em minus 0.4em\relax Citeseer, 2005.

\bibitem{tenorth2010knowrob}
M.~Tenorth, L.~Kunze, D.~Jain, and M.~Beetz, ``Knowrob-map-knowledge-linked
  semantic object maps,'' in \emph{Humanoid Robots (Humanoids), 2010 10th
  IEEE-RAS International Conference on}.\hskip 1em plus 0.5em minus 0.4em\relax
  IEEE, 2010, pp. 430--435.

\bibitem{kunze2010putting}
L.~Kunze, M.~Tenorth, and M.~Beetz, ``Putting people��s common sense into
  knowledge bases of household robots,'' in \emph{KI 2010: Advances in
  Artificial Intelligence}.\hskip 1em plus 0.5em minus 0.4em\relax Springer,
  2010, pp. 151--159.

\bibitem{tenorth2013knowrob}
M.~Tenorth and M.~Beetz, ``Knowrob: A knowledge processing infrastructure for
  cognition-enabled robots,'' \emph{The International Journal of Robotics
  Research}, vol.~32, no.~5, pp. 566--590, 2013.

\bibitem{chen2012toward}
X.~Chen, J.~Xie, J.~Ji, and Z.~Sui, ``Toward open knowledge enabling for
  human-robot interaction,'' \emph{Journal of Human-Robot Interaction}, vol.~1,
  no.~2, pp. 100--117, 2012.

\bibitem{xie2014understanding}
J.~Xie and X.~Chen, ``Understanding instructions on large scale for human-robot
  interaction,'' in \emph{Proceedings of the 2014 IEEE/WIC/ACM International
  Joint Conferences on Web Intelligence (WI) and Intelligent Agent Technologies
  (IAT)-Volume 03}.\hskip 1em plus 0.5em minus 0.4em\relax IEEE Computer
  Society, 2014, pp. 175--182.

\bibitem{xie2015multi}
J.~Xie, X.~Chen, and J.~Ji, ``Multi-mode natural language processing for
  human-robot interaction,'' in \emph{Web Intelligence}, vol.~13, no.~4.\hskip
  1em plus 0.5em minus 0.4em\relax IOS Press, 2015, pp. 267--278.

\end{thebibliography}

\end{document}